\documentclass[review,12pt]{elsarticle}

\usepackage[a4paper,margin=3cm,bottom=4.5cm]{geometry}
\usepackage{amssymb}
\usepackage{amsmath}
\usepackage{graphicx}%
\usepackage{multirow}%
\usepackage{amsfonts}%
\usepackage{mathrsfs}%
\usepackage{xcolor}%
\usepackage{textcomp}%
\usepackage{manyfoot}%
\usepackage{booktabs}%
\usepackage{algorithm}%
\usepackage{algorithmicx}%
\usepackage{algpseudocode}%
\usepackage{listings}%
\usepackage{subcaption}
\usepackage{makecell}
\RequirePackage{booktabs}
\usepackage{color}
\usepackage{url}
\usepackage{enumitem}
\usepackage{censor}
\usepackage{setspace}
\usepackage{bibspacing}
\usepackage{xurl}

\algnewcommand{\LeftAlignComment}[1]{\Statex \(\triangleright\) \textit{#1}}
\usepackage{tabularx,booktabs}
\usepackage{nicefrac}       
\usepackage{fancyhdr} 
\usepackage{pgfplots} 
\pgfplotsset{compat=1.18} 

\makeatletter
\def\ps@pprintTitle{%
  \let\@oddhead\@empty
  \let\@evenhead\@empty
  \def\@oddfoot{\reset@font\hfil\thepage\hfil}
  \let\@evenfoot\@oddfoot
}
\makeatother

\begin{document}

\begin{frontmatter}



\title{A Direct Classification Approach for Reliable Wind Ramp Event Forecasting under Severe Class Imbalance\tnoteref{pub_info}}
\tnotetext[pub_info]{Published in EPSR \url{https://doi.org/10.1016/j.epsr.2026.112951}}


\author[ULB]{Alejandro Morales-Hern\'andez \corref{mycorrespondingauthor}}

\cortext[mycorrespondingauthor]{Corresponding author}
\ead{alejandro.morales.hernandez@ulb.be}

\author[ULB,USannio]{Fabrizio De~Caro}\ead{fdecaro@unisannio.it}

\author[ULB]{Gian Marco Paldino}\ead{gian.marco.paldino@ulb.be}

\author[ULB]{Pascal Tribel}\ead{pascal.tribel@ulb.be}

\author[USannio]{Alfredo Vaccaro}\ead{vaccaro@unisannio.it}

\author[ULB,WELT]{Gianluca Bontempi}\ead{gianluca.bontempi@ulb.be}

\affiliation[ULB]{organization={Machine Learning Group, Universite Libre de Bruxelles},
            city={Brussels},
            postcode={1050}, 
            country={Belgium}}
\affiliation[WELT]{organization={WEL Research Institute},
            city={Wavre},
            postcode={1300}, 
            country={Belgium}}
            
\affiliation[USannio]{organization={University of Sannio},
            city={Benevento},
            postcode={82100}, 
            country={Italy}}

\begin{abstract}
Decision support systems are essential for maintaining grid stability in low-carbon power systems, such as wind power plants, by providing real-time alerts to control room operators regarding potential events, including Wind Power Ramp Events (WPREs). These early warnings enable the timely initiation of more detailed system stability assessments and preventive actions. However, forecasting these events is challenging due to the inherent class imbalance in WPRE datasets, where ramp events are less frequent (typically less than 15\% of observed events) compared to normal conditions. Ignoring this characteristic undermines the performance of conventional machine learning models, which often favor the majority class. This paper introduces a novel methodology for WPRE forecasting as a multivariate time series classification task and proposes a data preprocessing strategy that extracts features from recent power observations and masks unavailable ramp information, making it integrable with traditional real-time ramp identification tools. Particularly, the proposed methodology combines majority-class undersampling and ensemble learning to enhance wind ramp event forecasting under class imbalance. Numerical simulations conducted on a real-world dataset demonstrate the superiority of our approach, achieving over 85\% accuracy and 88\% weighted F1 score, outperforming benchmark classifiers.
\end{abstract}




\begin{keyword}


Wind Power Ramp Events \sep Class Imbalance \sep Machine Learning \sep Ensemble Methods \sep Time Series Classification
\end{keyword}

\end{frontmatter}

\section{Introduction}
\label{sec:introduction}

Decarbonization policies targeting greenhouse gas reduction are accelerating the transition to low-carbon power systems, where the effective integration of renewable energy sources is essential to meet regulatory targets. Wind energy stands out for its abundance and low environmental impact~\citep{ren2024robust}. However, wind power generators inherently exhibit low inertia and variable power output profiles, with sudden and significant changes over short periods defined as relevant Wind Power Ramp Events (WPREs)~\citep{han2020wind}. These events pose challenges to grid stability, increasing vulnerability to frequency deviations and potentially triggering emergency actions, such as load shedding or the disconnection of inverter-based generators, if not promptly addressed.~\citep{farooq2022frequency}. 

To mitigate the effects of high wind penetration, several strategies can be adopted, including synthetic inertia, integration of energy storage systems, deployment of grid-forming inverters, and flexible load management. This introduces the challenge of effectively anticipating critical events that may threaten the system and allocating flexibility resources accordingly~\citep{li2024prediction}. In this context, enhanced forecasting tools for WPREs can support intelligent decision-making systems, providing crucial assistance to control room operators during real-time operations~\citep{de2021daft}. The need for such tools is further emphasized by reports from system operators such as ERCOT~\cite{Wan2013} and AEMO~\cite{AEMO2020}.

These identification tools offer valuable insights into the anticipated operating conditions of the power system. Their development is driven by the need for system operators to receive timely, informed decision-making support, presented in a concise and interpretable format that can be quickly understood and acted upon~\cite{stankovic2022methods}. The outputs of such tools can either serve as direct inputs for operational decisions or act as triggers for more detailed, computationally intensive analyses such as dynamic security analysis~\cite{de2024role}.

While it may seem intuitive to use traditional wind power forecasting outputs to identify WPREs, dedicated tools are preferable due to fundamental differences between regression and classification in machine learning. Regression models predict continuous variables, such as wind power levels, whereas classification models are designed to identify discrete events like ramp occurrences or severities based on specific labeling rules. This distinction is crucial, as WPREs involve abrupt, high-magnitude changes that regression models may fail to capture effectively. Classification models, on the other hand, can be trained to recognize patterns and precursors of ramp events, enabling faster and more targeted operational responses.

WPREs often vary in duration, which complicates regression-based approaches that require multi-step predictions to capture the full event trajectory~\citep{pandit2025enhancing}. Classification approaches bypass this issue by directly identifying ramp events based on key time steps and power changes, without requiring precise modeling of their duration or full trajectory.

\subsection{State of the Art}\label{sec:related-works}

Forecasting WPREs is challenging due to the stochastic nature of wind and the complex interplay of meteorological factors that influence wind patterns~\citep{he2023trend}. In addition, WPREs are relatively rare compared to normal operating conditions, leading to highly imbalanced datasets where ramp events constitute a small fraction of the data~\citep{he2024wind}. This class imbalance constitutes a significant obstacle to traditional machine learning models, which may become biased towards the majority class and consequently exhibit poor ramp event detection performance~\citep{dhiman2021machine,chen2024survey}.

The state-of-the-art in WPRE forecasting and detection encompasses advanced signal processing techniques, deep learning models, trend analysis, optimization methods, and real-time data incorporation~\citep{gallego2015review}. Two general approaches can be identified to forecast WPREs: direct and indirect methods. Similar to conventional time series forecasting approaches, direct methods estimate the occurrence of ramp events by learning from past observations. By contrast, indirect methods involve an intermediate step, in which a numeric variable related to WPREs is first forecasted (e.g., generated power). Subsequently, an identification procedure is used to determine the ramp event type. A major limitation of indirect approaches lies in error propagation during WPRE identification, which depends heavily on the accuracy of the intermediate forecasts.

Deep learning models have been widely adopted to capture the complex temporal and spatial dependencies in wind power data, with most studies relying on indirect approaches for WPRE forecasting. \citet{li2024prediction} proposed a pipeline that combines denoising wind power generation data, feature extraction, and sequence modeling to perform indirect WPRE forecasting. Specifically, they forecast the wind power profile and subsequently apply a thresholding technique to identify ramp events. However, this regression-based approach does not directly address the multiclass nature of ramp events or the inherent class imbalance in classification datasets. Similarly, \citet{cui2023algorithm} proposed a wind power forecast model using historical wind power and Numerical Weather Prediction (NWP) data. This is followed by an improved Swinging Door Algorithm (ImDSDA) for WPRE detection and a clustering algorithm to assign events to four categories. As in the work of \citet{li2024prediction}, class imbalance is not explicitly considered, and the performance evaluation primarily focuses on binary ramp event detection rather than multiclass classification.

Trend analysis has received considerable attention in WPRE detection and forecasting. \citet{he2024wind} introduced ramp thresholds based on Value-at-Risk to quantify extreme wind power variations. Their approach extracts trend-related features and uses them as inputs to a classification model that directly forecasts WPRE categories, thereby avoiding intermediate variable prediction. The authors addressed class imbalance using SMOTE and undersampling techniques. However, SMOTE can generate unrealistic synthetic minority instances~\citep{sauglam2022novel}, while conventional undersampling may discard informative majority class instances, potentially degrading classification performance~\citep{liu2008exploratory}.

\citet{he2023trend} adopted the Swinging Door Algorithm (SDA) to detect and label up-trend and down-trend segments, combine adjacent segments, and revise their endpoints. Then, the authors include indicative variables (up, down, flat events) as inputs to a RNN to generate wind power forecasts, from which WPREs are subsequently detected. Binary classification metrics are computed to evaluate the detection capabilities of the method. The results demonstrate that incorporating trend-based indicative variables improves prediction accuracy and that RNNs can outperform alternative deep learning architectures.

\citet{li2020forecasting} proposed a direct method for forecasting wind power ramp events by defining these events based on significant changes in wind speed and installed capacity. The authors pre-process historical wind speed data and use an Event-Based K-means (EB-K) clustering algorithm to classify ramp events into multiple clusters, extracting key patterns that represent typical ramp event characteristics. These key patterns are then used to identify similar events in forecasted data through a rolling window approach and similarity measures. Although clustering implicitly performs a form of classification, the issue of class imbalance is not explicitly addressed.

\subsection{Research Gap and Proposed Work}

Despite progress in WPRE forecasting and detection, the majority of existing approaches remain indirect, first predicting wind power and then identifying ramp events based on significant forecasted variations. In contrast, direct methods forecast ramp events without relying on intermediate variables, enabling more timely and accurate alerts and thereby enhancing grid stability and operational response. Additionally, many existing methods neglect class imbalance, failing to account for the rarity of ramp events compared to normal conditions. This omission can bias models toward the majority class and degrade ramp event detection performance. Furthermore, these approaches often depend on large volumes of labeled historical data, limiting their applicability in real-time environments where data may be sparse, noisy, or lack timely event annotation.

Considering this, this paper proposes a novel methodology to enhance the forecasting accuracy of WPREs under class imbalance. Our method leverages an ensemble learning technique specifically designed for imbalanced datasets, namely \textit{EasyEnsemble}~\citep{liu2008exploratory}. By integrating this technique into the forecasting pipeline, we aim to improve the detection of rare ramp events while maintaining high overall prediction accuracy. The main contributions of this work are summarized as follows:

\begin{itemize}[nosep]

\item the development of a direct ramp event forecasting framework that explicitly addresses class imbalance, using only data acquired from remote terminal units and without relying on wind power forecasts or external meteorological inputs;

\item the integration of the proposed method with real-time ramp detection algorithms, such as SDA, thereby distinguishing it from conventional machine-learning-only approaches; 

\item the design of a novel data-preparation pipeline that extracts statistical features from ramp events, rather than focusing exclusively on generated power, and masks unlabeled data before model training, in accordance with the real-time availability of ramp event information;

\item the formulation of ramp event forecasting as a time series classification task that explicitly accounts for the temporal dependencies between power measurements and ramp occurrences, as well as the real-time constraints on ramp information availability.
\end{itemize}

The rest of the paper is organized as follows. Section~\ref{sec:problem} introduces the problem statement and details the data preparation process. Section~\ref{sec:approach} describes the proposed forecasting methodology, including the ensemble techniques employed. The validation strategy and experimental setup are outlined in Section~\ref{sec:validation_methodolgy}, while Section~\ref{sec:results} presents and analyzes the numerical results. Finally, Section~\ref{sec:conclusions} concludes the paper and outlines potential directions for future research.

\section{Problem definition and data preparation}
\label{sec:problem}

Let $x \in \mathbb{R^+}$ be a variable denoting generated power, observed over a discrete time scale within a period $t \in \{1, 2, \ldots, T\}$ where $T \in \mathbb{N}$ is the number of observations. Accordingly, the univariate time series of the generated power can be defined as a sequence of observations $\{x^{(t)}\}_{t=1}^{T} = \{x^{(1)}, x^{(2)}, \ldots, x^{(T)}\}$. Similarly, the sequence of ramp events is represented by a variable $r \in \mathcal{R}$ observed with the same time scale as $x$ to generate the univariate time series $\{r^{(t)}\}_{t=1}^{T} = \{r^{(1)}, r^{(2)}, \ldots, r^{(T)}\}$ which captures the temporal evolution of ramp events. Here, $\mathcal{R}$ is the discrete set of $C$ ramp events that may occur within the geographic area under analysis and that correspond to the classification target. Consequently, we define a bi-variate time series as the sequence $\{(x^{(t)}, r^{(t)})\}_{t=1}^{T} =\{ (x^{(1)}, r^{(1)}),(x^{(2)}, r^{(2)}), \dots, (x^{(T)}, r^{(T)}) \}$. At forecasting time $t$, we assume a model $\mathcal{F}$ has access to the last $l$ observations to forecast the type of ramp occurring at $t+h$, where $h$ corresponds to the forecasting horizon. Model $\mathcal{F}$ may also use additional features derived from these $l$ observations, which can contribute to the forecasting of the target.

Let us assume that $\mathcal{X} = \{(x^{(t)}, r^{(t)})\}_{t=1}^{T}$ is a dataset comprising the bi-variate time series of generated power and the sequence of ramp events. First, we need to extract a set $\mathcal{I}^l$ from $\mathcal{X}$ with $N$ training instances, each of them with the form $\{(x_i^{(t)}, r_i^{(t)})\}_{t=1}^{l}, \; l \in \mathbb{Z}^+, \; i \leq N$, and targets $\mathcal{Y}^h$ as $\{ r_i^{(l+h)}\}, \; i \leq N$; where $l$ represents the past steps used to forecast the type of ramp event occurring in the $h$ horizon. The instances are generated using a stride of $l+h$, ensuring that neither lagged observations nor target labels are shared across instances, which allows the subsequent use of stratified cross-validation without information leakage. For simplicity, we assume a multi-class and single-target classification problem ($h=1$). Second, each instance in $\mathcal{I}$ is flattened to obtain an $N \times 2l $ matrix. Subsequently, we mask the ramp information of observation $r_i^{(t)}, t<$ at t timestamps corresponding to those whose ramp type is unknown at time $t, t<l$, and augment the transformed matrix $\mathcal{I}$ with $m$ features extracted from the generated power sequence included in each instance. 

To emulate a real-time data stream, we consider the ramp type $r_i^{(t)}$ at time 
$t$ in instance $i$ as \textit{unknown} if the ramp state continues beyond the current subsequence window of length $l$. Formally, for $t<l$, the ramp is unknown at time $t$ if $r_i^{(t-1)} = r_i^{(t)} =  r_i^{(t+l+h)}$, where $h> 0$ accounts for the forecasting horizon, indicating the ramp sequence at time $t$ is ongoing and not yet finalized within the subsequence. In an online setting, the condition of an event being ongoing is determined incrementally by the SDA algorithm, which evaluates whether incoming observations continue the current ramp trend (line 4, Algorithm \ref{alg:pipeline}). In our offline experiments, we emulate this behavior using fully observed data: during instance construction, if the ramp label at the end of the subsequence remains unchanged in the next time step, the event is considered ongoing, and its type is masked. For example, for a subsequence of length $l=4$ with ramp labels at times $t-3$, $t-2$, $t-1$, and $t$ all equal to an increasing ramp, if the label at $t+1$ is also increasing, the event is ongoing, and the last observation is masked (see Supplementary materials for a visual example). By masking unavailable or ongoing ramps, the supervised learning task treats the occurrence of an unknown event until time $t$ as a feature. These masked values are represented by a constant outside the domain and should not be interpreted as missing data that can be imputed. Rather, the masking reflects the fundamental uncertainty of the event, which cannot be reconstructed until it has concluded.

\section{Proposed approach}
\label{sec:approach}

Algorithm~\ref{alg:pipeline} formalizes the proposed workflow for real-time ramp event forecasting. We assume that a system continuously receives generated power measurements and that, upon the arrival of each new observation, the SDA algorithm evaluates whether it conforms to the current power trend. If the new value marks a change in the trend, an SDA-based decision rule is applied to identify the ramp event that was active up to that moment, and the new observation is masked as \textit{unknown} and treated as the initial observation of a new event. Subsequently, the data preparation step described in Section \ref{sec:problem} transforms the available information into a suitable input representation for the machine learning model, which then forecasts the type of ramp event expected to occur next. The ML model is trained offline using historical records from the study area, and we do not consider a feedback loop from the incoming data to retrain the model. 

Notice that an SDA-based classification is only applicable when there is a change in the observed trend so far; in other cases, this approach is unable to provide a classification. Hence, the need for algorithms such as our proposal. The SDA-based classification depends on the ramp definition assumed in the analysis. In general, the definition of WPRE must include at least three features: duration, magnitude, and direction~\cite{cui2021algorithm}. We adopt a WPRE definition that characterizes a ramp event as a variation of the wind power output exceeding a predefined threshold within a given time interval (typically 20\% of the installed capacity within at most 4 hours). This definition is formally expressed in Equation~\ref{eq:ramp_def}:

\begin{equation}
\label{eq:ramp_def}
     \left\{\begin{matrix}
max \, P^{[t: t+\Delta t]}  - min \, P^{[t: t+\Delta t]} \geq P_{\text{Capacity}} \times \omega \\
\Delta t \leq 4 \; hours
\end{matrix}\right.
\end{equation}

\noindent where $P^{[t, t+\Delta t]}$ denotes a certain power value in the interval defined by $t$ and $\Delta t$; $P_{\text{Capacity}}$ is the installed capacity of the wind farm, and $\omega$ is the threshold assumed to consider that a ramp is observed.

\begin{algorithm}
\caption{\textit{Real-time forecasting of ramp events}}\label{alg:pipeline}
\begin{algorithmic}[1]
\Require $\mathcal{F}$: A ML model trained with historical records
\Require $i$: Index of last known event
\Require $\Omega$: set of thresholds for SDA-based classification

\State $\overset{\circ}{\mathbf{\mathcal{X}}}, \overset{\circ}{\mathbf{\mathcal{Y}}} \leftarrow \left[ \; \right ]$ \Comment{\textit{Historical records}}

\While{$j=1, \dots, \infty$}
    \State $\mathbf{\mathcal{X}}^{(j)} \leftarrow $ Read available data from the data stream

    \If{$\mathbf{\mathcal{X}}^{(j)}$ does not belong to the current trend} \Comment{SDA algorithm}
        \State $R^{(j)} \leftarrow$ Compute ramp ratio for $(\overset{\circ}{\mathbf{\mathcal{X}}}^{[i+1:j]}, \overset{\circ}{\mathbf{\mathcal{Y}}}^{[i+1:j]})$  \Comment{\textit{Starting event}}
        \State $c \leftarrow$ Event type based on $R^{(j)}$ and $\Omega$  
        \State $\overset{\circ}{\mathbf{\mathcal{Y}}} \leftarrow \overset{\circ}{\mathbf{\mathcal{Y}}} \cup c$ \Comment{\textit{Update historical records with $c$}}
    \EndIf
    
    \State $\overset{\circ}{\mathbf{\mathcal{X}}} \leftarrow \overset{\circ}{\mathbf{\mathcal{X}}} \cup \mathbf{\mathcal{X}}^{(j)}$

    \State $\mathcal{I}^l \leftarrow$ Data preparation for $(\overset{\circ}{\mathbf{\mathcal{X}}}^{[i:j]}, \overset{\circ}{\mathbf{\mathcal{Y}}}^{[i:j]})$
    \State \textbf{output} $\mathcal{F}(\mathcal{I}^l)$ \Comment{\textit{Forecast with ML model}}
\EndWhile

\end{algorithmic}
\end{algorithm}

The following features are extracted during the data preparation phase, and only from the generated power observed in the training instances defined by $l$, instead of the complete time series:

\begin{itemize}[nosep]
    \item Maximum, mean, median, minimum, variance
    \item Signal distance between the start and end of the sample
    \item Shannon entropy
    \item Interquartile range
    \item Lempel-Ziv’s (LZ) complexity index, normalized by the sample size ($l$).
    \item Mean and median absolute deviation from the mean
    \item Mean absolute difference between observations $ \left< x^{(t)}, x^{(t+1)} \right>_{t=1}^{l-1}$ 
    \item Mean difference between observations $ \left< x^{(t)}, x^{(t+1)} \right>_{t=1}^{l-1}$ 
    \item Median absolute difference between observations $ \left< x^{(t)}, x^{(t+1)} \right>_{t=1}^{l-1}$ 
    \item Median difference between observations $ \left< x^{(t)}, x^{(t+1)} \right>_{t=1}^{l-1}$ 
    \item Number of negative and positive turning points of the signal
    \item Fractal dimension using Petrosian's algorithm.
    \item Root mean square of the time series
    \item Slope of a linear equation fitted to the sample
    \item Last known event type: Used to capture the probabilistic dependence between past and future events, and available from the second recorded event onward or through expert knowledge.
\end{itemize}

\begin{algorithm}
\caption{\textit{EasyEnsemble}}\label{alg:easy_ensemble}
\begin{algorithmic}[1]
\Require $\mathcal{P}$: A set of non-majority class examples
\Require $\mathcal{M}$: A set of majority class examples
\Require $\mathcal{L}$: The number of subsets $\mathcal{L}$ to sample from $M$
\Require $S$: The number of weak classifiers in an AdaBoost ensemble $\mathcal{H}$

\While{$i=1, \dots, \mathcal{L}$} 
    
    \State Randomly sample a subset $N_i$ from $\mathcal{M}$, $ \left | \mathcal{N}_i \right | \approx \left | \mathcal{P} \right |$
    
    \State Learn $\mathcal{H}_i$ using $\mathcal{P}$ and $\mathcal{N}_i$. $\mathcal{H}_i$ is an AdaBoost ensemble with $S$ weak classifiers $h_{i,j}$ and corresponding weights $\alpha_{i,j}, j=1,\dots,S$. The ensemble's threshold is $\theta_i$,

    \[
    \mathcal{F}_i = sgn\left ( \sum_{j=1}^{S} \alpha_{i,j}h_{i,j}(x)-\theta_i\right )
    \]
    
\EndWhile

\State \Return an ensemble
\[
    \mathcal{F} = sgn\left ( \sum_{i=1}^{L}\sum_{j=1}^{S} \alpha_{i,j}h_{i,j}(x)-\sum_{i=1}^{L}\theta_i \right )
    \]
\end{algorithmic}
\end{algorithm}

The classification task of identifying ramp events is highly unbalanced since non-significant events or normal conditions occur more frequently than ramp events. Therefore, it is likely that an ML model that ignores the class imbalance will suggest the most common one, increasing the amount of False Negative predictions (events that can be considered a ramp but are classified as normal conditions instead). Undersampling is a popular strategy for dealing with class imbalance problems since it uses only a subset of the majority class and is therefore very efficient. The key flaw is that many majority-class examples are overlooked. To tackle this problem, we propose to use \textit{EasyEnsemble} \citep{liu2008exploratory} (Algorithm~\ref{alg:easy_ensemble}) in the pipeline presented in Algorithm~\ref{alg:pipeline}. Several studies \citep{paldino2024role,rezvani2024methods,siregar2024enhancing,alsorory2024boosting} have demonstrated the superior performance of \textit{EasyEnsemble} compared to other algorithms, showcasing its effectiveness and reliability in achieving better results under class imbalance.

\textit{EasyEnsemble} is a straightforward approach to exploit further the majority-class instances ignored by undersampling. This method independently samples several subsets $N_1, N_2, \dots, N_L$ from the set $N$ that represent the set of instances with the majority class. For each subset $N_i; \; 1 \leq i \leq L$, a classifier $H_i$ is trained using $N_i$ and the instances of the other classes. AdaBoost \citep{schapire1999brief} is used to train the classifier $H_i$, which in turn is composed of $S$ weak classifiers. Finally, instead of counting votes from ${H_i}; i=1,\dots, L$, the algorithm collects all the weak classifiers $h_{i,j} \; (i=1,\dots, L,  j = 1, \dots, S)$ and form an ensemble with them. By combining Adaboost learners trained on different data subsets, \textit{EasyEnsemble} has benefited from the combination of boosting and a bagging-like strategy with balanced class distribution.

\section{Validation methodology}
\label{sec:validation_methodolgy}

As a result of previous collaborations with the Italian Transmission System Operator, we adopted a dataset provided by the University of Sannio with 28 800 samples, which roughly represent 80 days. This dataset represents the aggregated wind power generated by several wind turbines in a wind farm, with measurements recorded every 15 minutes. To ensure the privacy of the data source, no information about the sampling period or geolocation is shared, and the data was transformed using the MinMax scaling method. The type of ramp event at each time interval was provided (see supplementary materials for more details), together with the generated power. The former was obtained assuming 10\% (for a 3-class problem) and 20\% (for a 5-class problem) of a wind farm with an installed capacity of 669 MW.

\begin{figure}[!htbp]
\centering

\includegraphics[width=\textwidth]{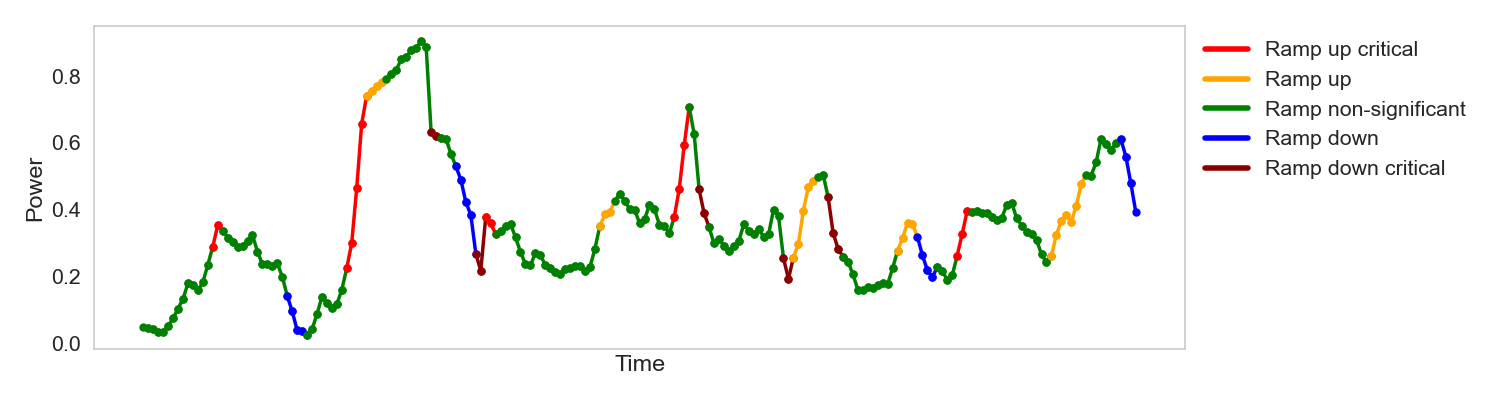}

\caption{Sample of the wind power time series adopted in this study. Five different events are highlighted with different colors to indicate the five different classes considered in the forecasting task. }
\label{fig:time_series}
\end{figure}

Figure~\ref{fig:time_series} shows a subsample of the power time series and the sequence of ramp events. Five types of ramp events are considered as classes in the current study: ``critical ramp-UP events'' (class \textit{ramp-UP critical}), ``ramp-UP events'' (class \textit{ramp-UP}), ``non-significant ramp events'' (class \textit{no ramp}), ``ramp-DOWN events'' (class \textit{ramp-DOWN}), and ``critical ramp-DOWN events'' (class \textit{ramp-DOWN critical}). To the best of our knowledge, our study is the first to analyze this number of classes since most of the papers published on this topic assume 2 or 3 event types. 

\begin{figure}[!htbp]
\centering

\includegraphics[width=\textwidth]{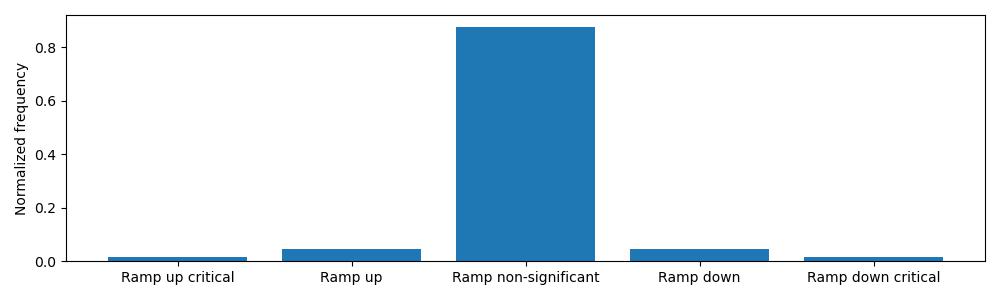}

\caption{Probability of ramp occurrence in the dataset}
\label{fig:umbalance}
\end{figure}

Figure~\ref{fig:umbalance} shows the probability of occurrence of each ramp event type in the dataset. As expected, non-significant ramp events or normal conditions are more frequent than the other types (more than 88\% of the events were normal). Therefore, the algorithm proposed in the forecasting pipeline of our approach will sample several subsets of the training instances whose class is assigned to be a non-significant ramp event. Based on the number of classes, we considered two different experimental settings: a first experiment with the original class definition to train the ML model and another experiment where the number of classes is reduced by grouping the classes \textit{ramp-DOWN} and \textit{ramp-DOWN critical} into the new class \textit{``ramp-DOWN*''}, and classes \textit{ramp-UP} and \textit{ramp-UP critical} into the new class \textit{``ramp-UP*''}.

In addition to our proposal, we analyze the performance of five other algorithms, each of which handles the class imbalance differently. They are as follows:

\begin{itemize}[nosep]
    \item Random forest classifier (RF) with the class weight adjusted to consider the balance of the classes

    \item Balanced Random forest classifier (BRF)\citep{chen2004using} that handles the class unbalance by drawing a bootstrap sample from the minority class and sampling with replacement the same number of samples from the majority class.

    \item RUSBoost classifier (RB)\citep{seiffert2009rusboost} which integrates random under-sampling in the learning of AdaBoost.

    \item Balanced Bagging classifier (BB) that balances the training set during training time using random undersampling.

    \item Long Short-Term Memory network for classification (LSTMc). Class-specific weights, calculated based on the number of instances per class, were incorporated into the cross-entropy loss function during training to address class imbalance in the dataset.

\end{itemize}

Lastly, we included another classifier (Pr) in the benchmark as a naive baseline, which does not consider the class balance to forecast the type of the next event. Equation~\ref{eq:probabilistic_classifier} formalizes this baseline as the probability of an event $r^{(t+1)}$ occurring in the future given an event $r^{(t)}$. Notice that if we assume that $P(r | r^{(t)}) = \frac{1}{\left| \mathcal{R} \right|}$, then each class has an equal chance of occurring, meaning that the algorithm will randomly forecast the type of event occurring next. 

\begin{equation}
    \label{eq:probabilistic_classifier}
    \begin{aligned}
        \mathcal{F}_{\text{baseline}}(r^{(t)}) & = r^{(t+1)} \sim S[P(r | r^{(t)})]; \; r, r^{(t)} \in \mathcal{R} =  \{ \text{``ramp-UP critical''}, \\ \text{``ramp-UP''}, 
        & \text{``no ramp''}, \text{``ramp-DOWN''}, \textit{``ramp-DOWN critical''} \}
    \end{aligned}
\end{equation}

As the last known event type cannot be inferred for the first event in the dataset, only samples from the second event onward were considered in the experimental evaluation. We split the dataset using a hold-out approach (80\% for training and 20\% for testing purposes) after the original dataset was transformed following the pre-processing steps discussed in section~\ref{sec:problem}. Then, stratified cross-validation with $k=5$ was used on the training set to optimize the hyperparameters of each model. Lastly, each model was retrained to use the best hyperparameter configuration, and the overall performance was evaluated using the test set. The library HyperOpt was considered to perform hyperparameter optimization (See the Supplementary Materials for more details) using the weighted F1 score as the optimization objective.

Unlike binary classification, the classes in multi-class classification cannot be directly represented as positive (the class of interest) or negative (the other class) of a value. Hence, we cannot compute True Positive (TP), True Negative (TN), False Positive (FP), and False Negative (FN) values straightforwardly. Instead, we need to calculate TP, TN, FP, and FN for each class ($TP^c$, $TN^c$, $FP^c$, and $FN^c$, respectively). As a result, the metrics used to assess the performance of ML models in multi-class classification are calculated accordingly. As for the performance metric, we use the model accuracy (Eq.~\ref{eq:accuracy}), the balanced accuracy defined as the average recall on each class (Eq.~\ref{eq:balanced_accuracy}), Cohen's kappa score (Eq.~\ref{eq:kappa}), and the weighted F1 score (Eq.~\ref{eq:weighted_F1Score}). The weights $w$ in Eq. \eqref{eq:weighted_F1Score} are determined according to the number of true instances for each class. In addition, we report the training and testing time of each forecasting model.

\begin{equation}
\label{eq:accuracy}
     \text{Accuracy} = \tfrac{TN+TP}{TN+TP+FN+FP}
\end{equation}

\begin{equation}
\label{eq:balanced_accuracy}
\begin{aligned}
     \text{Balanced accuracy} &= \frac{1}{\left| \Omega \right|}\sum_{c \in \Omega} \text{Recall}^c \\
     \text{Recall}^c & = \tfrac{TP^{c}}{TP^{c} + FN^{c}}
\end{aligned}
\end{equation}

\begin{equation}
\label{eq:kappa}
\begin{aligned}
     \text{Cohen's kappa score} & = \tfrac{\text{Accuracy} - p_e}{1-p_e} \\
     p_e & =\sum_{c \in \Omega}\tfrac{(TN^c_{\text{truth}}+FP^c_{\text{truth}})*(TN^c_{\text{estimate}} + FN^c_{\text{estimate}})}{TN+TP+FN+FP}
\end{aligned}
\end{equation}

\begin{equation}
\label{eq:weighted_F1Score}
\begin{aligned}
     \text{Weighted F1 score} & = \sum_{c \in \Omega} w \tfrac{2*\text{Precision}^c*\text{Recall}^c}{\text{Precision}^c + \text{Recall}^c} \\
     \text{Precision}^c & = \tfrac{TP^{c}}{TP^{c} + FP^{c}} 
\end{aligned}
\end{equation}

\section{Experimental results}
\label{sec:results}

Table \ref{tab:lags_analysis} shows an analysis of the influence of the number of lags on the model's performance and the number of classes in the classification task. The parameter was varied in the discrete set $l=\{4, 8, 12\}$ which corresponds to 1, 2, and 3 hours respectively, given the sampling frequency of the data. A reduction in model performance was generally observed when $l$ increases, meaning that a few observations closer to the current time $t$ are enough to forecast the next event at time $t+1$. Therefore, the results discussed from here on are those obtained for $l=4$ unless noted otherwise. As expected, the imbalance in the original dataset, measured through the Shannon entropy (Table \ref{tab:lags_analysis}), is higher than when the number of classes is reduced. Even when reducing the number of classes does not ensure a better class balance, the impact of the class imbalance is clear in the algorithm's performance through the algorithm performance (best results are observed for 3 classes).

\begin{table}[!hbtp]
\centering
\caption{Performance of \textit{EasyEnsemble} with different lag values and number of classes. The best results are highlighted for each metric}
\label{tab:lags_analysis}
\begin{tabular}{llllll}
\multicolumn{6}{c}{Reducing the number of classes (from 5 to 3 classes)} \\ \hline
\multicolumn{1}{c}{\textbf{Lag}} & \makecell{\textbf{Shannon} \\ \textbf{entropy (\%)}} & \multicolumn{1}{c}{\textbf{Accuracy}} & \multicolumn{1}{c}{\makecell{\textbf{Balanced} \\ \textbf{accuracy}}} & \multicolumn{1}{c}{\makecell{\textbf{Kappa }\\\textbf{score}}} & \multicolumn{1}{c}{\makecell{\textbf{Weighted } \\ \textbf{F1 score}}} \\ \hline
4 (1 hour) & 41.95 & \textbf{0.907} & \textbf{0.915} & \textbf{0.679} & \textbf{0.92} \\
8 (2 hours) & 41.955 &  0.875 & 0.611 & 0.49 & 0.882 \\
12 (3 hours) & \textbf{41.959} & 0.853 & 0.699 & 0.5 & 0.874 \\ \hline 
\multicolumn{6}{c}{Using the original class definition (5 classes)} \\ \hline 
\multicolumn{1}{c}{\textbf{Lag}} & \makecell{\textbf{Shannon} \\ \textbf{entropy (\%)}} & \multicolumn{1}{c}{\textbf{Accuracy}} & \multicolumn{1}{c}{\makecell{\textbf{Balanced} \\ \textbf{accuracy}}} & \multicolumn{1}{c}{\makecell{\textbf{Kappa }\\\textbf{score}}} & \multicolumn{1}{c}{\makecell{\textbf{Weighted } \\ \textbf{F1 score}}} \\ \hline
4 (1 hour) & 33.084 & 0.862 & \textbf{0.68} & \textbf{0.542} & \textbf{0.883} \\
8 (2 hours) & 33.088 & 0.863 & 0.458 & 0.445 & 0.872 \\
12 (3 hours) & \textbf{33.091} & \textbf{0.869} & 0.439 & 0.417 & 0.866 \\ \hline
\end{tabular}
\end{table}

\begin{figure}[!hbtp]
    \centering
    \begin{subfigure}[b]{.48\textwidth}
        \includegraphics[width=\textwidth]{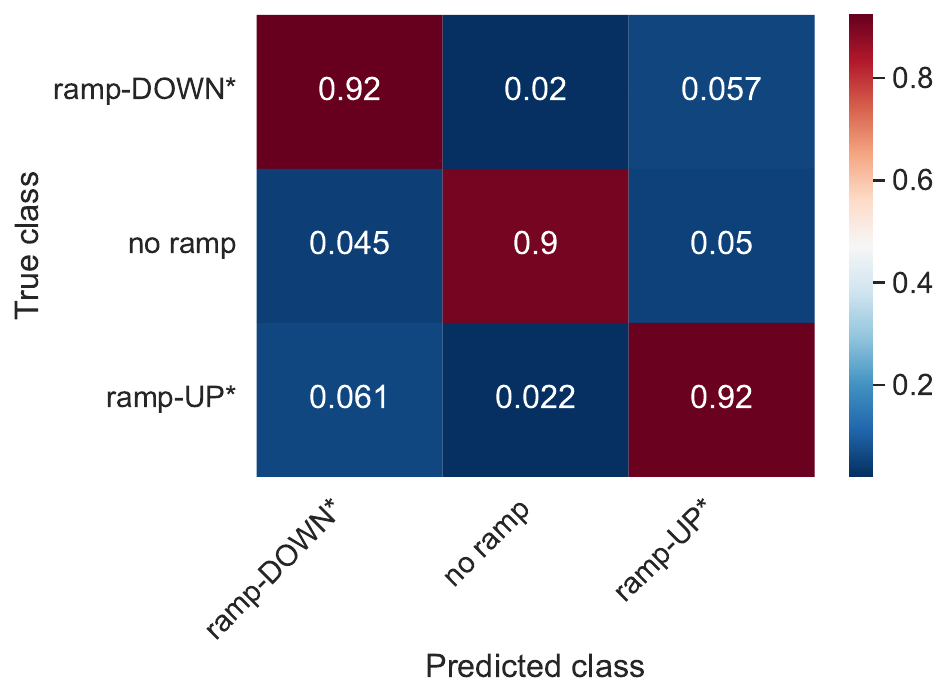}
        \caption{Reducing the number of classes (from 5 to 3 classes)}
        \label{fig:cm_3classes}
    \end{subfigure}
    \hfill
    \begin{subfigure}[b]{.48\textwidth}
        \includegraphics[width=\textwidth]{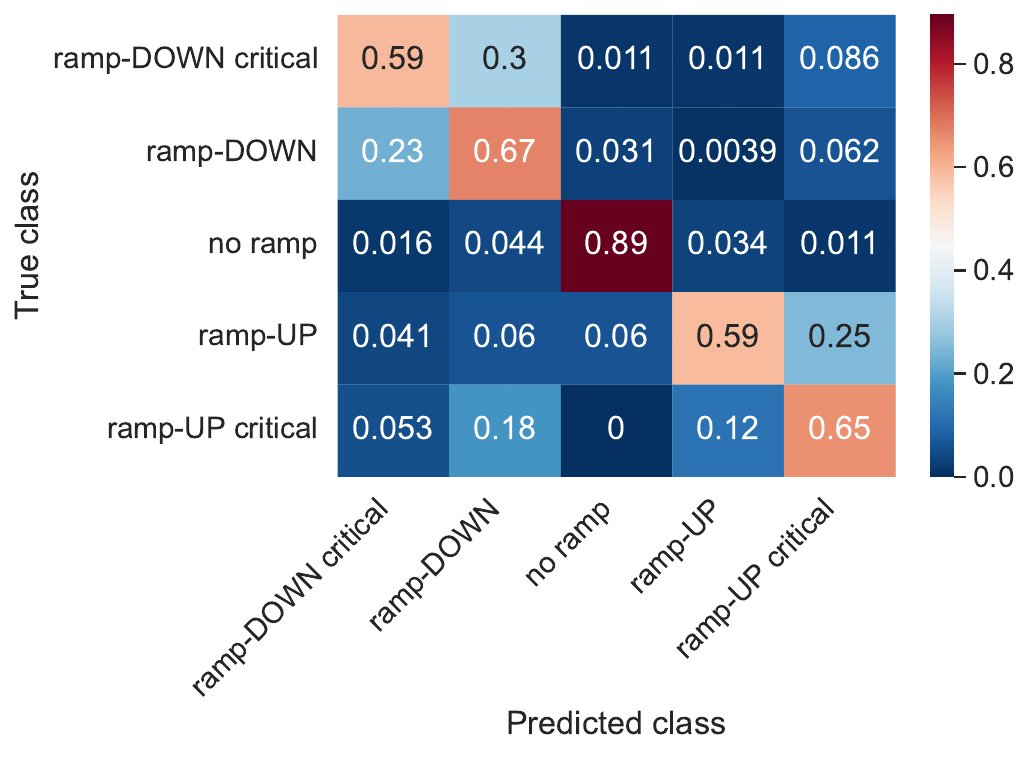}
        \caption{Original class definition (5 classes)}
        \label{fig:cm_5classes}
    \end{subfigure}
    \caption{Confusion matrix of the classification results using \textit{EasyEnsemble} with $l=4$ and (a) 3 classes and (b) 5 classes. The values were normalized over the true conditions (i.e. rows) }
    \label{fig:cm_EasyEnsemble}
\end{figure}

The performance of \textit{EasyEnsemble} is summarized in a confusion matrix (Figure \ref{fig:cm_EasyEnsemble}), comparing the predictions and true values. The values are normalized over the true conditions to make the classification results easier to interpret. As highlighted in Table \ref{tab:lags_analysis}, \textit{EasyEnsemble} can accurately estimate the ramp event type, especially when the classification problem is defined with 3 classes. In this case, the algorithm can distinguish the minority classes (\textit{``ramp-DOWN''} and \textit{``ramp-UP''}) from the majority class, which is the main challenge in imbalance classification. For example, 92\%, 90\%, and 92\% of the instances observed as \textit{``ramp-DOWN*''}, ``\textit{normal}'', and \textit{``ramp-UP*''} respectively, were classified correctly (Figure \ref{fig:cm_3classes}). However, it is more difficult for the algorithm to distinguish the ``ramp-DOWN'' and ``ramp-UP'' instances from the ``ramp-DOWN critical'' and ``ramp-UP critical'' respectively when the original class definition is used (Figure \ref{fig:cm_5classes}). For instance, we can see that the algorithm tends to incorrectly classify the ``ramp-DOWN critical'' instances as ``ramp-DOWN'' (59\% versus 30\%). Still, it is more confident to identify that a ``ramp-DOWN critical'' is not any other event. Similarly, 18\% of the ``ramp UP critical'' instances were misclassified as ``ramp-DOWN'' (12\% as ``ramp UP'') (last row in the confusion matrix of Figure \ref{fig:cm_5classes}). This is not surprising given the class imbalance of the dataset. 

Table \ref{tab:benchmark_analysis} shows that \textit{EasyEnsemble} consistently outperforms the other approaches when the classification challenge is to distinguish between three ramp events. Balanced Bagging was the second-best algorithm in the benchmark, and the performance observed with the original class definition was close to the ones obtained with \textit{EasyEnsemble} w.r.t. the accuracy and the weighted F1 score. \textit{EasyEnsemble} profited from the combination of boosting and a bagging-like method with balanced class distribution. Both \textit{EasyEnsemble} and Balanced Bagging attempt to use balanced bootstrap samples. However, the former uses the samples to produce boosted ensembles (thus reducing the classification error after training a new tree), and the latter uses them to train decision trees at random. It is known that boosting mainly reduces bias while bagging mainly reduces variance. We hypothesize that the improvement achieved in the forecasting is the direct result of the integration of the best features of the two methodologies.

\begin{table}[!hbtp]
\centering
\caption{Results for forecasting the ramp event's occurrence using $l = 4$ (1 hour). The best results are highlighted for each metric}
\label{tab:benchmark_analysis}

\begin{tabular}{lllllll}
\multicolumn{7}{c}{Reducing the number of classes (from 5 to 3 classes)} \\ \hline
\multicolumn{1}{c}{\textbf{Algorithms}} & \multicolumn{1}{c}{\textbf{Acc.}} & \multicolumn{1}{c}{\makecell{\textbf{Balanced} \\ \textbf{accuracy}}} & \multicolumn{1}{c}{\makecell{\textbf{Kappa}\\\textbf{score}}} & \multicolumn{1}{c}{\makecell{\textbf{Weighted } \\ \textbf{F1 score}}} & \multicolumn{1}{c}{\makecell{\textbf{Training} \\ \textbf{time (s)}}} & \multicolumn{1}{c}{\makecell{\textbf{Test} \\ \textbf{time (s)}}}\\ \hline
Our proposal & \textbf{0.907} & \textbf{0.915} & \textbf{0.679} & \textbf{0.917} & 2710.68 & 19.4 \\
BB & 0.849 & 0.679 & 0.473 & 0.867 & 0.16 & $<0.01$ \\
BRF & 0.819 & 0.727 & 0.437 & 0.846 & 0.51 & $<0.01$ \\
RF & 0.777 & 0.724 & 0.385 & 0.816 & 1.21 & $<0.2$\\
RB & 0.877 & 0.727 & 0.54 & 0.887 & 0.65 & $<0.01$ \\
LSTMc & 0.483 & 0.563 & 0.137 & 0.579 & 79.69 & 3.17 \\
Pr & 0.859 & 0.519 & 0.386 & 0.861 & 0.13 & $0.1$ \\ \hline

\multicolumn{7}{c}{Using the original class definition (5 classes)} \\ \hline
\multicolumn{1}{c}{\textbf{Algorithms}} & \multicolumn{1}{c}{\textbf{Acc.}} & \multicolumn{1}{c}{\makecell{\textbf{Balanced} \\ \textbf{accuracy}}} & \multicolumn{1}{c}{\makecell{\textbf{Kappa}\\\textbf{score}}} & \multicolumn{1}{c}{\makecell{\textbf{Weighted } \\ \textbf{F1 score}}}  & \multicolumn{1}{c}{\makecell{\textbf{Training} \\ \textbf{time (s)}}} & \multicolumn{1}{c}{\makecell{\textbf{Test} \\ \textbf{time (s)}}} \\ \hline
Our proposal & \textbf{0.862} & \textbf{0.68} & \textbf{0.542} & \textbf{0.883} & 640.14 & 18.48\\
BB & 0.861 & 0.482 & 0.447 & 0.872 & 1.14 & $<0.1$ \\
BRF & 0.846 & 0.445 & 0.379 & 0.855 & 0.5 & $<0.1$ \\
RF & 0.77 & 0.464 & 0.312 & 0.814 & 1.88 & $0.02$ \\
RV & 0.831 & 0.458 & 0.39 & 0.849 & 0.25 & $0.01$\\
LSTMc & 0.44 & 0.351 & 0.094 & 0.563& 86.14 & 3.42\\
Pr & 0.648 & 0.264 & 0.08 & 0.711 & 0.09 & $0.1$ \\ \hline
\end{tabular}
\end{table}

\begin{figure}[!htbp]
\centering

\includegraphics[width=\textwidth]{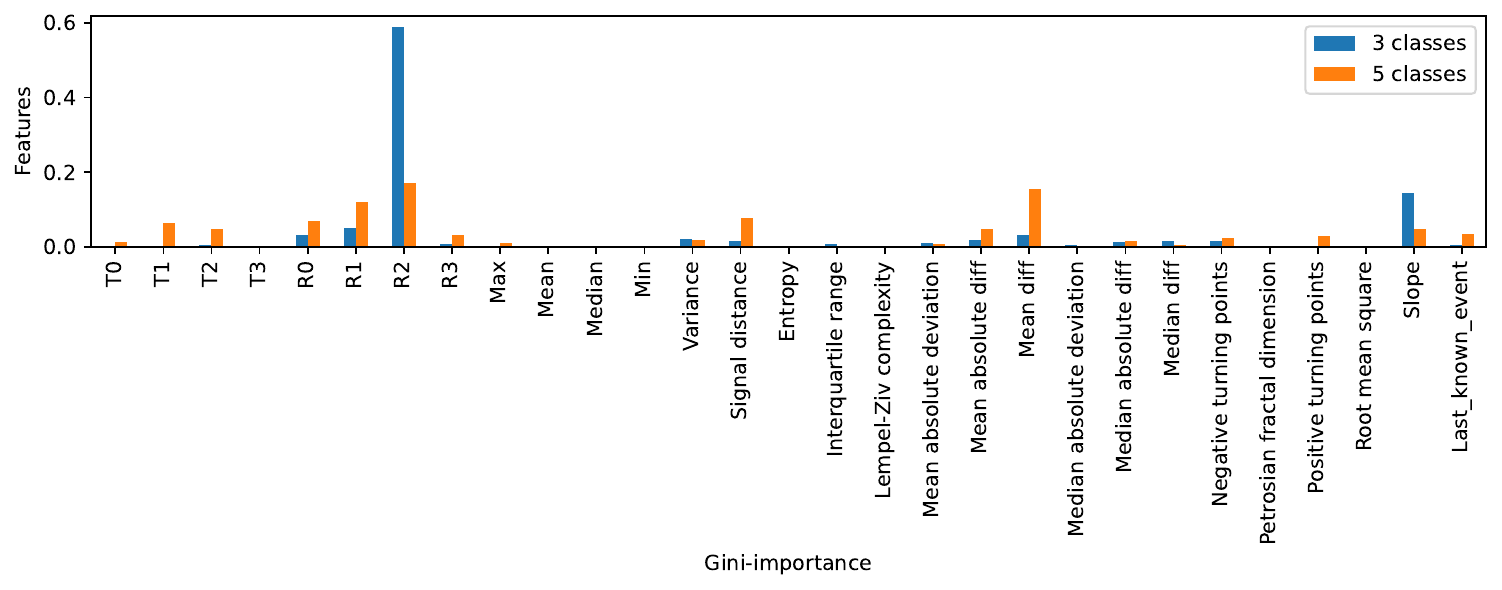}

\caption{Feature relevance in \textit{EasyEnsemble} computed through the Gini Importance or Mean Decrease in Impurity (MDI)}
\label{fig:feature_importance}
\end{figure}

We use the Gini Importance or Mean Decrease in Impurity (MDI) metric to determine the relevance of each feature in the forecasting task. The importance of a feature is computed as the (normalized) total reduction of the criterion brought by that feature\footnote{\url{https://scikit-learn.org/dev/modules/generated/sklearn.tree.DecisionTreeClassifier.html\#sklearn.tree.DecisionTreeClassifier.feature_importances\_}}. Therefore, the higher the MDI the relevant the feature. Figure \ref{fig:feature_importance} shows that the feature ``R2'' (the third feature representing the event type at time $t=3$ from a sequence of size 4) was consistently the most relevant feature, both using the original class definition and a small number of classes. Besides ``R2'', the slope of the time series (feature ``Slope'') is the second most relevant feature for \textit{EasyEnsemble} when three classes are considered. The influence of other static features on the classification is more evident when the original class definition is used. Here, the mean difference between observations and the signal distance (feature ``mean diff'' and ``signal distance'' respectively) were almost as relevant as the type of previous ramp events (features ``R2'', ``R1'', and ``R0''). In general, these results highlight the additional value of including the new features proposed for the ramp event categorization task.

\section{Conclusions}
\label{sec:conclusions}

This paper addresses WPRE forecasting under class imbalance by formulating it as a multivariate time series classification task. We introduce a data pre-processing strategy that extracts informative features from recent power observations and masks unavailable ramp information to emulate real-time operating conditions, thereby ensuring practical applicability. To mitigate class imbalance, we employ \textit{EasyEnsemble}, which effectively leverages majority class examples without aggressive undersampling, thus preserving training efficiency. Experiments on real-world wind power data show substantial improvements in forecasting performance, particularly for minority ramp event classes, with \textit{EasyEnsemble} achieving the highest accuracy and weighted F1 scores in a three-class classification setting. Feature importance analysis highlights the relevance of previous event types and statistical descriptors, such as slope and mean difference, in enhancing predictive performance. Overall, these findings contribute to improved grid stability and risk management by increasing the predictability of WPREs.

Despite the promising results obtained, several directions for future research remain open. First, the incorporation of additional features, such as meteorological data or numerical weather predictions, could further strengthen the model's predictive capabilities by capturing more complex interactions between wind dynamics and ramp events. Second, exploring online learning methodologies could allow the model to adapt to changing conditions in real-time. In this study, the model is trained offline and remains static after deployment. Third, investigating advanced deep learning architectures, such as Transformer models with self-attention mechanisms, could provide further improvements. Although these models have demonstrated success in capturing long-term dependencies in time series data, their application for class-imbalanced problems remains largely underexplored. 

\section*{Credit author statement}
\textbf{Alejandro Morales-Hernández}: Conceptualization, Methodology, Software, Validation, Writing - Original Draft, Visualization. \textbf{Fabrizio de Caro}: Conceptualization, Methodology, Software, Data curation, Writing - Original Draft, Supervision. \textbf{Gian Marco Paldino}: Software, Writing - Original Draft. \textbf{Pascal Tribel}: Software, Writing - Review \& Editing. \textbf{Alfredo Vaccaro}: Supervision, Writing - Review \& Editing. \textbf{Gianluca Bontempi}: Supervision, Writing - Review \& Editing

\section*{Declaration of competing interest}

The authors declare that they have no known competing financial interests or personal relationships that could have appeared to influence the work reported in this paper

\section*{Data statement}

The data that support the findings of this study are available from the corresponding author upon reasonable request.

\section*{Acknowledgement}
 This research work has been supported by the Service Public de Wallonie Recherche under grant nr 2010235–ARIAC by DigitalWallonia4.ai., and by the Joint R\&D TORRES Project (2022-RDIR-59b) funded by ``R\'{e}gion de Bruxelles-Capitale - Innoviris''. The research also benefited from the support of the Walloon Region to Pr. G. Bontempi  as part of the funding for the FRFS‑WEL-T strategic axis. 

\singlespacing
 \bibliographystyle{elsarticle-num-names} 
 \bibliography{sn-bibliography}

\end{document}


\begin{frontmatter}

\title{Supplementary material  for ``\textit{A Direct Classification Approach for Reliable Wind Ramp Event Forecasting under Severe Class Imbalance}''}

\end{frontmatter}


\section{Data pre-processing pipeline and real-time application}

In a truly online setting, the condition of an event being ongoing is determined incrementally by the event detection module, in our case, the SDA algorithm, which assesses whether incoming observations continue the current ramp trend. In an offline experimental setting, however, we must emulate this real-time behavior using fully observed data.

Concretely, during instance construction, we determine whether the ramp observed at time $t$ is finalized within the subsequence window of length $l$ (i.e., $t+l$ represents the current window of analysis) or continues beyond it (in our case, $t+l+1$, representing the future). This is achieved by comparing the ramp labels within the window $r(t+l)$  with the label at $r(t+l+1)$, which represents the event outcome that would only become known in the future in an online scenario. If the ramp label remains unchanged, the event at time $t+l$ (or at least the last observations in this window) is considered ongoing, and its type is masked, reflecting the fact that its final classification is not yet available at that time.

As an illustrative example, consider a subsequence of length $l=4$ where the ramp labels at times $t-3$, $t-2$, $t-1$, and $t$ are all equal to an increasing ramp. If the label at $t+1$ is also an increasing ramp, this indicates that the event extracted within the window has not yet terminated. In an online setting, SDA would still identify the ramp as ongoing, and its final type would be unavailable. The masking procedure reproduces this situation by preventing the model from accessing information that would only be known after the event ends.

\begin{figure}[!htbp]
\centering
\begin{subfigure}[b]{\textwidth}
\centering
\includegraphics[width=13cm]{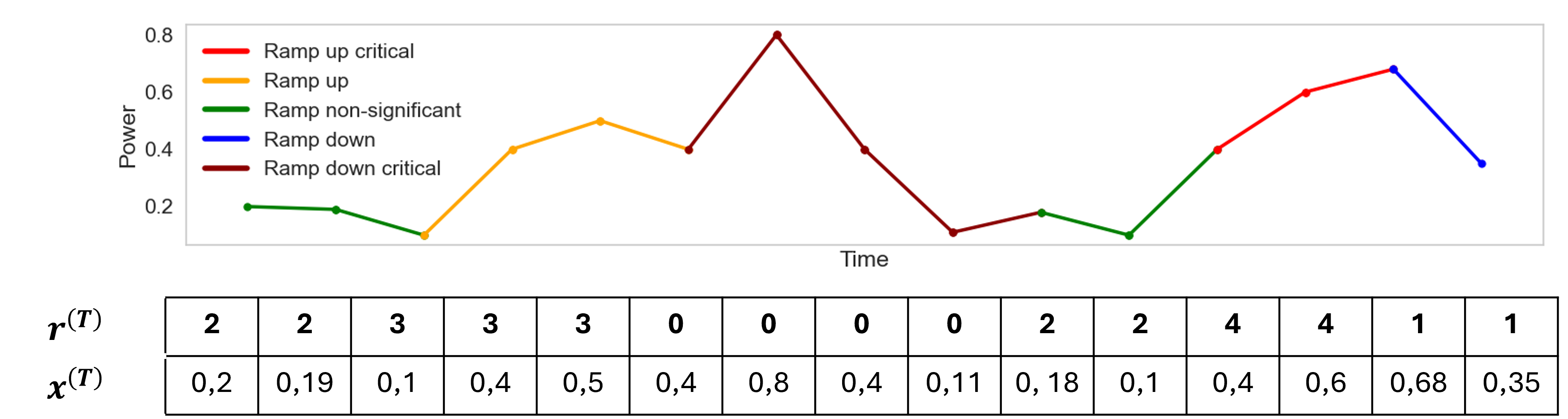}
\caption{Original bivariate time series with $x \in \mathbb{R^+}$ and $r \in \mathcal{R}, |\mathcal{R}|=C$\vspace{1mm}}
\label{fig:preprocessingA}
\end{subfigure}
\vfill

\begin{subfigure}[b]{\textwidth}
\centering
\includegraphics[width=6cm]{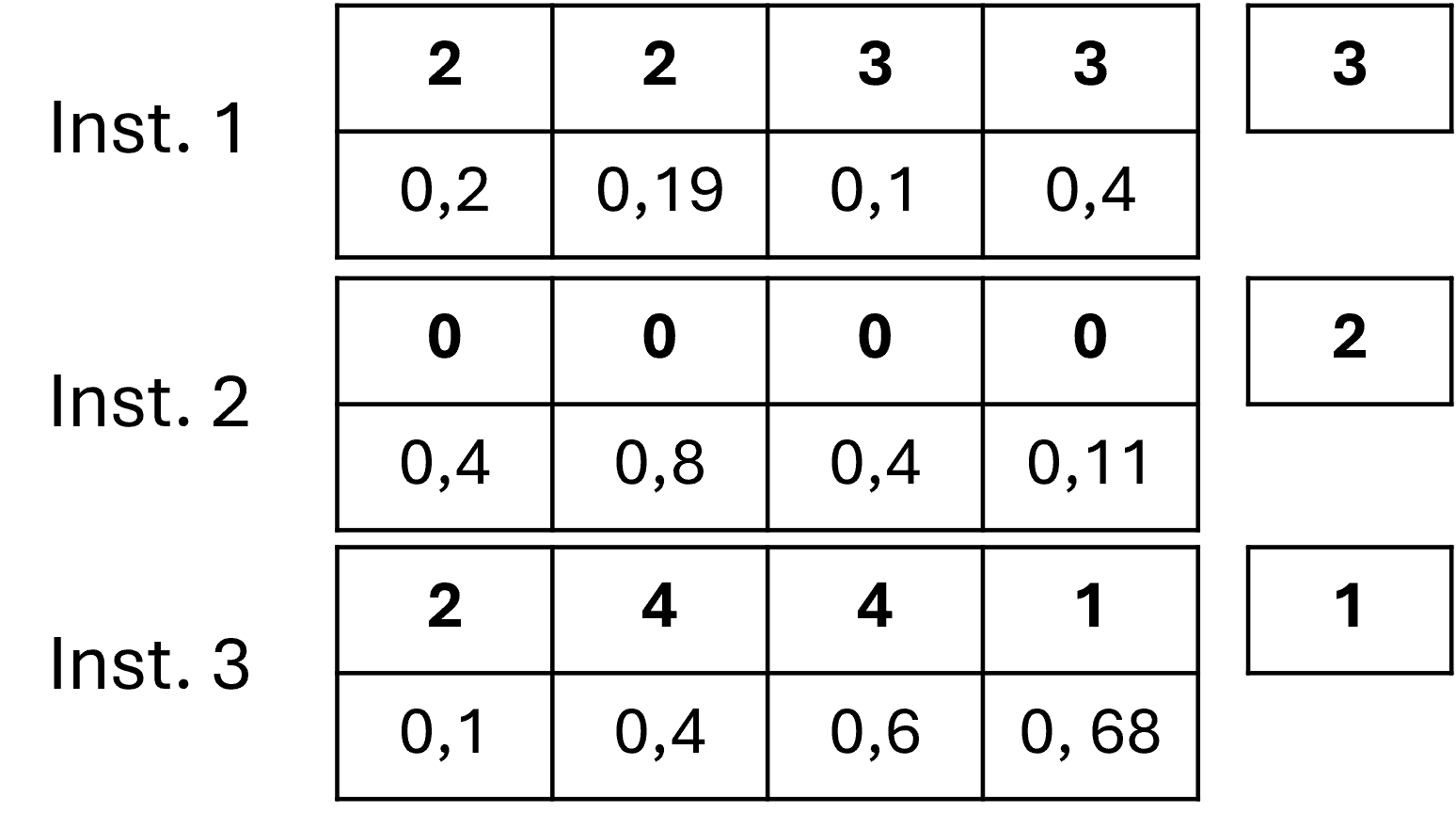}
\caption{Moving window of size $l=4$ to extract training instances to predict the type of event at time $l+h, h=1$}
\label{fig:preprocessingB}
\end{subfigure}
\vfill

\begin{subfigure}[b]{\textwidth}
\centering
\includegraphics[width=9cm]{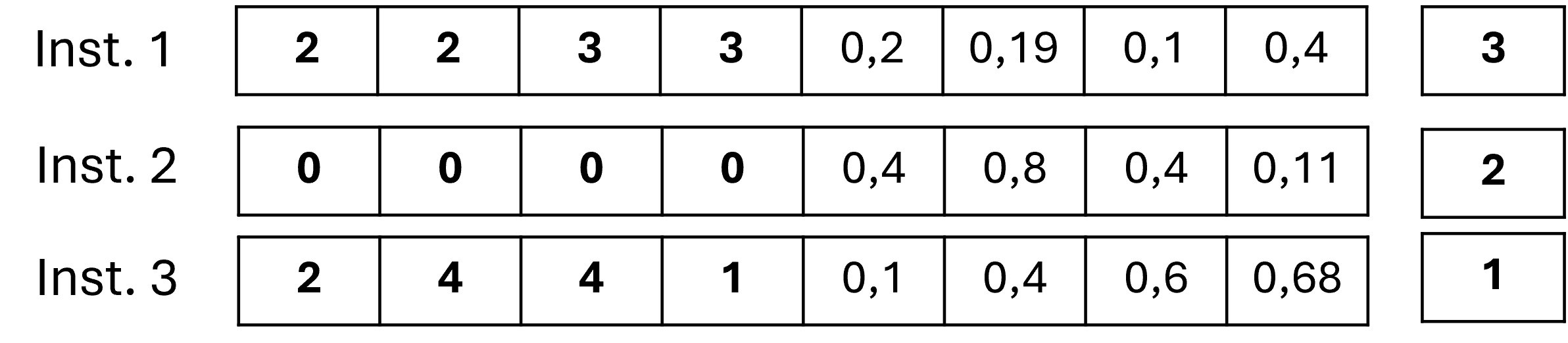}
\caption{Flattening}
\label{fig:preprocessingC}
\end{subfigure}

\begin{subfigure}[b]{\textwidth}
\centering
\includegraphics[width=12cm]{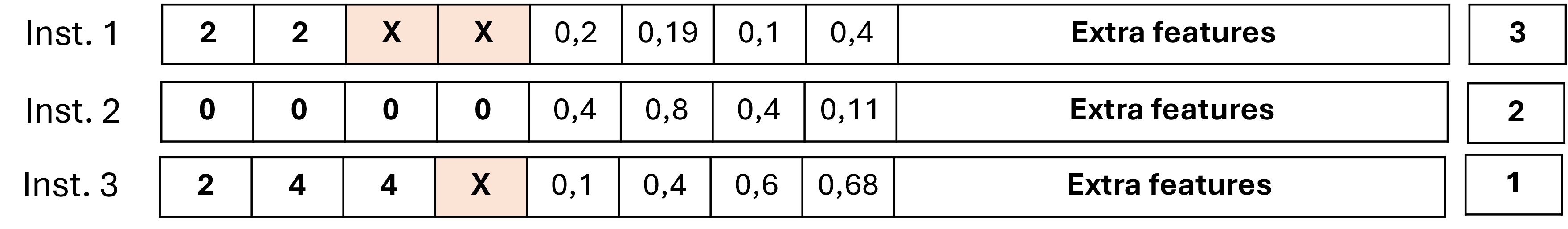}
\caption{Masking and feature extraction}
\label{fig:preprocessingD}
\end{subfigure}

\caption{Data pre-processing using $l=4$ and four events ($C=4$). (a) The original bivariate time series $\{(x^{(t)}, r^{(t)})\}_{t=1}^{T}$, with rows as variables and columns as timestamps. (b) Selection of training instances and targets by applying a moving window on the time series (with a stride of 5). (c) Each instance is flattened, and (d) the information regarding the ramp type is masked depending on its availability, and additional features are extracted from the generated power window.}
\label{fig:preprocessing}
\end{figure}

Figure~\ref{fig:preprocessing} summarizes the pre-processing pipeline through a simple example of a time series of length 15. 

\begin{figure}[!htbp]
\centering
\includegraphics[width=13cm]{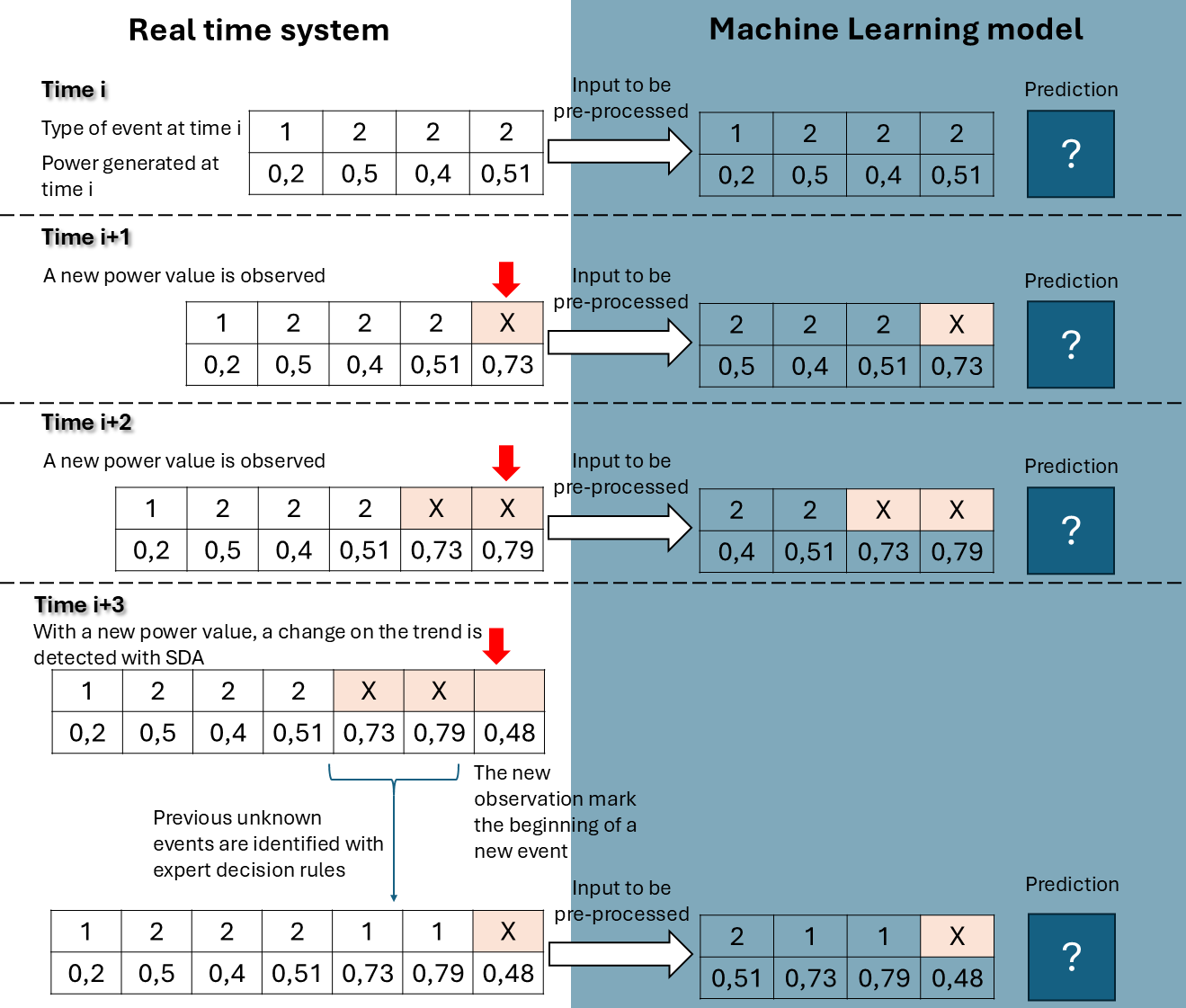}

\caption{Real-time processing of the bivariate time series for ramp event forecasting}
\label{fig:real_time_forecasting}
\end{figure}

Figure~\ref{fig:real_time_forecasting} presents an additional toy example illustrating how a continuous data stream is processed to forecast ramp events, using $l=4$. As detailed in Algorithm 1, each time a new observation is received, that is, when a generated power value is recorded, SDA evaluates whether it is consistent with the previously observed trend. If the trend is maintained, the ongoing event is assumed to be still in progress, and the ramp event type is masked in the instance provided to the machine learning model (see the transition from \textit{time i} to \textit{time i+1}). This procedure is repeated for each subsequent observation until SDA detects a change in the trend. Once a change is identified, an expert decision rule (see section s\ref{sec:sda_assignment}) is applied to classify the event that has just concluded, and the new observation is treated as the start of a new, yet unknown, event (see the transition from \textit{time i+2} to \textit{time i+3}). 

\section{SDA-based assignment of ramp event type}
\label{sec:sda_assignment}

Once SDA detects a change in the observed trend, an expert-defined decision rule is applied to assign the ramp event type.

\begin{equation}
    \left\{\begin{matrix}
\text{ramp down critical} & ; Rr \leq -s_2 \\
\text{ramp down} & ; Rr > -s_2 \; \text{and}\; Rr \leq -s_1 \\
\text{normal conditions} & ; Rr > -s_1 \; \text{and}\; Rr < s_1 \\
\text{ramp up} & ; Rr \geq s_1 \; \text{and}\; Rr < s_2 \\
\text{ramp upcritical} & ; Rr \geq s_2 \\
\end{matrix}\right. 
\end{equation}

\noindent where $Rr=\frac{\Delta P^{[t, t+\Delta t]}}{\Delta t}$ is the ramp ratio (MW/min) in the interval defined by $t$ and $\Delta t$, $s_1=\frac{0.1P_{\text{Capacity}}}{60}$ and $s_2=\frac{0.2P_{\text{Capacity}}}{60}$ are the thresholds assumed to consider that a specific ramp type is observed, and $P_{\text{Capacity}}=669 MW$ is the installed capacity of the wind farm.

\section{Hyperparameter optimization}
\label{sec:hyperparameters}

Table~\ref{tab:ee_hyperparameters} shows the hyperparameters optimized for \textit{EasyEnsemble}. The library HyperOpt was considered to perform hyperparameter optimization (75 iterations) using the weighted F1 score as the optimization objective.

\begin{table}[!hbtp]
\centering
\caption{Hyperparameter configuration of \textit{EasyEnsemble}}
\label{tab:ee_hyperparameters}
\begin{tabular}{lll|ll}
\hline
\multicolumn{1}{c}{\multirow{2}{*}{\textbf{Hyperparameter}}} &  \multicolumn{1}{c}{\multirow{2}{*}{\textbf{Type}}} &  \multicolumn{1}{c|}{\multirow{2}{*}{\textbf{Domain}}} &  \multicolumn{2}{c}{\textbf{Optimal value}} \\ \cmidrule{4-5} 
\multicolumn{1}{c}{} & \multicolumn{1}{c}{} &  \multicolumn{1}{c|}{} &  \multicolumn{1}{c}{\textbf{3 classes}} &  \multicolumn{1}{c}{\textbf{5 classes}} \\ \hline
Max depth             & Int.         & {[}1, 10{]}                          & 7         & 4           \\
Estimators ($\mathcal{L}, S$)           & Int.         & {[}1, 200{]}                         & 199        & 90          \\
Criterion             & Cat. & \makecell{{[}`log\_loss', `gini',\\ `entropy'{]}} & `entropy' & `entropy' \\
Min samples to split  & Cont.   & (0, 1{]}                             & 0.159   & 0.327    \\
Min samples in a leaf & Cont.   & (0, 1)                               & 0.03   & 0.059     \\
Learning rate         & Cont.  & {[}1, 5{]}                           & 1.035    & 1.006      \\ \hline
\end{tabular}
\end{table}

The configuration of the other algorithms included in the benchmark is as follows:

\textbf{Balanced Bagging (3 classes)}: `max depth': 29, `estimators': 8, `criterion': `entropy', `min samples to split': 0.419, `min samples in a leaf': 0.268. \textbf{(5 classes)}: `max depth': 14, `estimators': 145, `criterion': `entropy', `min samples to split': 0.415, `min samples in a leaf': 0.439  

\textbf{Balanced random Forest (3 classes)}: `max depth': 5, `estimators': 122, `criterion': `entropy', `sampling strategy': `not minority', `min samples to split': 0.636, `min samples in a leaf': 0.244, `class weight': `balanced\_subsample'. \textbf{(5 classes)}:  `max depth': 21, `estimators': 117, `criterion': `entropy', `sampling strategy': `not minority', `min samples to split': 0.308, `min samples in a leaf': 0.218, `class weight': `balanced\_subsample'. 

\textbf{Random Forest (3 classes)}: `max depth': 29, `estimators': 111, `criterion': `log\_loss', `min samples to split': 0.59, `min samples in a leaf': 0.104, `class weight': `balanced'. \textbf{(5 classes)}:  `max depth': 29, `estimators': 145, `criterion': `log\_loss', `min samples to split': 0.561, `min samples in a leaf': 0.104, `class weight': `balanced'. 

\textbf{RUSBoost (3 classes)}: `max depth': 20, `estimators': 31, `learning rate': $10^{1.128}$,  `criterion': `entropy', `min samples to split': 0.497, `min samples in a leaf': 0.13. \textbf{(5 classes)}: `max depth': 34, `estimators': 146, `learning rate': $10^{1.544}$,  `criterion': `gini', `min samples to split': 0.527, `min samples in a leaf': 0.102.

\textbf{LSTMc (3 classes)}: `hidden dimensions': 143, `learning rate': $10^{2.485}$, `layers': 2, `weight decay': 0.2614. \textbf{(5 classes)}: `hidden dimensions': 102, `learning rate': $10^{2.471}$, `layers': 3, `weight decay': 0.2076.